\documentclass[11pt,a4paper]{article}

\pdfoutput=1
\usepackage{color}
\usepackage[pdftex]{graphicx}
\usepackage[a4paper, lmargin=0.1666\paperwidth, rmargin=0.1666\paperwidth, tmargin=0.1111\paperheight, bmargin=0.1111\paperheight]{geometry} 
\usepackage{algorithm}
\usepackage{algpseudocode}

\usepackage{caption}
\DeclareCaptionFormat{hrformat}{#1#2#3\hrulefill}
\usepackage{fancyvrb}
\usepackage{url}

\usepackage{setspace} 
\usepackage{etoolbox}
\AtBeginEnvironment{quote}{\par\singlespacing\small}

\newcommand{\bsl}[0]{\ensuremath{\textrm\textbackslash}}

\title{
  Functional Benchmarks for Robust Evaluation of Reasoning Performance, and the
  {\em Reasoning Gap}
}
\author{Saurabh Srivastava\footnote{Corresponding author. Please send all correspondence to saurabh@consequent.ai.},\\
Annarose M B,
Anto P V,
Shashank Menon, Ajay Sukumar,\\
Adwaith Samod T,
Alan Philipose,
Stevin Prince, and
Sooraj Thomas\\\\
Consequent AI
}

\date{}

\begin{document}
\maketitle

\section{Abstract} We propose a framework for robust evaluation of reasoning
capabilities of language models, using {\em functional variants} of benchmarks.
Models that solve a reasoning test should exhibit no difference in performance
over the static version of a problem compared to a snapshot of the functional
variant. We have rewritten the relevant fragment of the MATH benchmark into its
functional variant MATH(), with functionalization of other benchmarks to
follow.  When evaluating current state-of-the-art models over snapshots of
MATH(), we find a {\em reasoning gap}---the percentage difference between the
static and functional accuracies. We find reasoning gaps from 58.35\% to
80.31\% among the state-of-the-art closed and open weights models that perform
well on static benchmarks, with the caveat that the gaps are likely to be
smaller with more sophisticated prompting strategies. Here we show that models
which anecdotally have good reasoning performance over real-world tasks, have
quantifiable lower gaps, motivating the open problem of building {\em ``gap
0''} models. Code for evaluation and new evaluation datasets, three MATH()
snapshots, are publicly available at
\url{https://github.com/consequentai/fneval/}.

\section{Introduction}

Accurately evaluating reasoning performance is critical to improving large
language models (LLMs) beyond their current capabilities. The current standard
for benchmarking fails to accurately measure reasoning of LLMs. An accurate
test should consist of posing a question whose answer can be {\em derived} under the
assumption of axioms for that domain. Reasoning systems (or humans) should be
able to conclude the answer despite not having seen the question before. An
ideal system would provide not only the answer but also each step of the
derivation.
Explicit elucidation of steps is ideal, but not critical. State-of-the-art
language models exhibit {\em some} reasoning capabilities, but accurate evaluation is
lagging. In particular, the capabilities of LLMs  may be overestimated because
they are tested using benchmarks that are sufficient for language understanding
(textual question, answer) but may not accurately measure reasoning.

When testing generalized problem-solving, the aim is to assess whether the
tested system can effectively answer questions it has not encountered before.
This is the opposite objective when compared to knowledge retrieval. 
Finding novel questions is challenging when the model has been trained on a
snapshot of the internet. Even if datasets exist that are not currently on the
internet, it is unreasonable to expect them to remain so, or that we will
continue to develop novel datasets. Contamination-testing by checking for
$k$-contiguous token sequences fails to account for various other forms in which
leakage might occur, especially when the benchmark is a set of static text
question-answers (``static QAs''). These may include i) accidental leakage
occurring from paraphrasing or compressed representations in the training data
or fragments of semantically similar statements; ii) not accounting for the
provenance of data used in every stage of training or fine-tuning, which
becomes increasingly difficult as more models are trained from existing
high-quality starting weights; iii) sourcing synthetic training data from an
existing model which could be contaminated, especially if it is closed-source
and closed-data, iv) overfitting through preference tuning, which may in the
extreme case cause reinforcement feedback allowing the model to evolve to
discover correct static answers.

Checking against a static QA is only an indirect test of reasoning. Text
questions and answers work well for humans as producing correct answers
implies likely correct stepwise reasoning.  The same does not apply to model
outputs.

We propose an alternative approach that alleviates concerns with static QA
testing, but also works seamlessly with existing evaluation harnesses. We
propose rewriting each static QA's reasoning into code, whose inputs can be
varied to create infinite versions that follow the same reasoning pattern, but
the steps needed in each would be different. We call this the {\em functional
variant} of the benchmark, and running each of those code fragments will create
a {\em snapshot} that is a drop-in replacement into the existing static
evaluation harness.  The accuracy on these replacements, with the static
variant included, might be lower than that on the static variant alone, which
we quantify as the {\em reasoning gap}.

For domains where this can be done, employing functional variants would improve
on static QA evaluation. We would convert a static benchmark $Bench$
(e.g., MATH or GSM8K or HumanEval) into their functional variant
$Bench(rngs)$. The functional variants (denoted here as QA($rngs$)) take as
inputs a set of deterministic typed random generators $rngs$ which output a
pseudorandom typed value, e.g., natural int, positive or negative float,
fractions, string, positive evens or odds, primes. The $rngs$ when seeded make
the instantiation deterministic, generating a snapshot of each test
(denoted here as QA*) and of the entire benchmark.

Each QA* text pair in this snapshot is different from the
original static QA. However, answering QA*  requires the same
reasoning used in the static QA. To have an unbiased accurate baseline, we
created the functionalization manually by rewriting the
reasoning of each test into code. Functionalization---a one-time
time-consuming task---results in infinite snapshot instantiations
to test against.

Of the popular benchmarks, functionalization is easy for the subcategories of
math (MATH~\cite{math} and GSM8K~\cite{gsm8k}), and code
(HumanEval~\cite{humaneval}, MBPP~\cite{mbpp}, and Natural2Code~\cite{gemini}).
The categories of commonsense reasoning (HellaSwag~\cite{hellaswag},
Winogrande~\cite{winogrande}, PIQA~\cite{piqa}, SIQA~\cite{siqa},
OpenbookQA~\cite{openbookqa}, ARC-Easy/Challenge~\cite{arc-easy-challenge},
CommonsenseQA~\cite{commonsenseQA}), world knowledge
(NaturalQuestions~\cite{naturalquestions}, TriviaQA~\cite{triviaQA}), and
reading comprehension (BoolQ~\cite{boolq}, QuAC~\cite{quac}, DROP~\cite{drop})
test English understanding and thus are not good candidates for
functionalization.  Subparts of the aggregated benchmarks (MMLU~\cite{mmlu},
HELM~\cite{helm}, BBH~\cite{big-bench-hard}, and AGI Eval~\cite{agi-eval}) could
be functionalized.

As of this work, we have functionalized 41.2\% of the MATH benchmark, with the
subset chosen so as to fully cover the static QA tests solved by 4 closed-weight and 9
open-weight state-of-the-art (SOTA) models. This allows us to provide complete metrics
over all of MATH for these models, assuming a simple few-shot prompting strategy.
We find a reasoning gap of 58.35\% to 80.31\% across these models.
The value of the gap when
using optimized prompting such as chain-of-thought
(CoT~\cite{chain-of-thought}), tree-of-thought (ToT~\cite{tree-of-thought}),
chain-of-code (CoC~\cite{chain-of-code}), amongst others, could be lower and we
will resolve that open question when we have built the100\% functionalized MATH().

In future work, we will present
functionalized variants of GSM8K, HumanEval, and MBPP.

\subsection{Contributions}
This work presents the first steps towards reasoning evaluation with functional benchmarks:
\begin{itemize}
  \item {\em Framework for evaluation for models that
    have seen the text of the entire internet:}
    We present a proposal for a long-term solution to the problem of evaluating
    models that are increasingly trained on all written knowledge. The proposed
    framework allows a benchmark to be instantiated to infinite snapshots, with
    a probabilistic guarantee that a new snapshot will not have existed before.
    For a static QA in the benchmark whose individual test captures a specific
    reasoning process, each snapshot QA* will follow the same process, but
    the test will be new. For a model to perform well on a set of
    snapshots, it will need to be able to perform step-by-step derivations to
    reach an answer, rather than simply recalling a static final answer.

  \item {\em Functional version of MATH, an important benchmark
    for reasoning evaluation, and its publicly available snapshots:}
    Our framework is designed to allow the reworking of existing benchmarks on
    which models are already being tested without modification. We are
    in the process of functionalizing the entire MATH benchmark.
    The specific functionalization code will never
    be publicly accessible. Instead three snapshots will be released every quarter.
    The \{Oct, Nov, Dec\}-2023 snapshots are available at
    \url{https://github.com/consequentai/fneval/} along with the evaluation code.
    The repository will be kept up to date with the last quarter's snapshots.

  \item {\em Robust
    evaluations that are resistant to contamination:}
    There are infinite snapshots of a functional benchmark, with the reasoning
    involved in each remaining  identical. If a model consistently solves a
    problem across multiple snapshots (e.g., three snapshots) then there is a
    high likelihood it does so with proper reasoning, because it should be
    fairly difficult to get the answer right across many snapshots by chance.

  \item {\em Reasoning gap:} We define the {\em reasoning
    gap} as the quantitative measure of the difference in accuracy when
    testing against static vs functional variants.
    The gap is a measure of reasoning vs memorization, with gap-0 being true
    reasoning and gap-100 being full memorization. The community should aim to
    build models that have the highest accuracy while maintaining a gap close
    to 0. Minimizing the gap may serve as a training optimization objective
    that could lead to models that perform better at generalized reasoning.

  \item {\em Open problem of training gap-0 models:}
    Here we show that even the current best models have large reasoning gaps,
    with the caveat that the reasoning gap is likely to be smaller with more
    sophisticated prompting strategies. Our current datasets, training,
    fine-tuning, and inference strategies might be non-optimal for training
    gap-0 models. We leave it as an open problem to train gap-0 models.

\end{itemize}

\section{Illustrative Examples}
We motivate the need for more accurate assessment of the reasoning capabilities
of LLMs by examining specific cases in the MATH benchmark.
Below and after, we use monotype font to indicate verbatim text from the
benchmark. In particular, the benchmark problems include LaTeX formatting, which
we leave as is to accurately display the model input.
For all problems below, these static QAs are solved by current LLMs---i.e.,
the inferred answer matches the ground truth answer shown.

\paragraph{Case 1 - Simple arithmetic}
\begin{verbatim}
  Question: Compute $\\left(-\\sqrt{5321}\\right)^2$.
  Answer: 5321

  Question: Round 15.49999999 to the nearest whole number.
  Answer: 15
\end{verbatim}
Solving problems such as these is in line with known capabilities of current models.
We would expect such problems to be solvable in their
functionalized form as well. We find this to be the case.

\paragraph{Case 2 - Non-trivial arithmetic or use of numerical properties}
\begin{verbatim}
  Question: $20!$ has 19 digits, the last 18 of which are
            432902008176640000. What is the first digit?
  Answer: 2
\end{verbatim}
The solution presented in the benchmark (for human reasoning) exploits a
property of divibility by $9$ to arrive at the answer. A model equipped with a
calculator could explicitly compute $20!$ and arrive at the same answer.
We do not expect the functionalized forms to be solvable by open weight models,
for whom we can confirm no tools are used in the inference pipeline. We find
this to be the case.

\paragraph{Case 3 - Undergraduate-level mathematical insights and skills}
\begin{verbatim}
  Question: What value of $x$ will give the minimum value
            for $x^2- 14x + 3$?
  Answer: 7

  Question: The equations $x^3 + 5x^2 + px + q = 0$ and
            $x^3 + 7x^2 + px + r = 0$ have two roots in common.
            If the third root of each equation is represented
            by $x_1$ and $x_2$ respectively, compute the ordered
            pair $(x_1,x_2).$
  Answer: (-5,-7)
\end{verbatim}
The first problem requires understanding when a quadratic equation is
minimal.  The second problem requires taking the difference of two polynomials,
observing a cancellation, and using properties of roots of quadratic and cubic
polynomials to arrive at the answer.  Neither of these are trivial, and we only
expect their functionalized forms to be solvable by top models that are capable
of symbolic manipulation.  We find that to be the case.

\paragraph{Case 4 - Graduate-level mathematics}
\begin{verbatim}
  Question: Euler discovered that the polynomial $p(n) = n^2 - n + 41$
            yields prime numbers for many small positive integer values
            of $n$. What is the smallest positive integer $n$ for which
            $p(n)$ and $p(n+1)$ share a common factor greater than $1$?
  Answer: 41
\end{verbatim}
While this problem shows solvable in its static QA form, the
graduate-level reasoning required should be outside the planning capabilities of
current models. We find that the functional forms are able distinguish such
problems and no models are able to solve the functional snapshots.

\paragraph{Case 5 - Problems where the ground truth answer is problematic.}
\begin{verbatim}
  Question: Find all solutions to the equation $\\sqrt{3x+6}=x+2$. If
            there are multiple solutions, order them from least to
            greatest, separated by comma(s).
  Answer: 1

  Question: Two concentric circular regions have radii of 1 inch and
            10 inches. What is the area, in square inches, outside
            the smaller region, but inside the larger region? Express
            your answer in terms of $\\pi$.
  Answer: 99\\pi \\text\{ square inches\}
\end{verbatim}
For the first question, the ground truth that matches the question is ``$-2,1$''
and the stated answer is incorrect. 
For the second question, there is extra verbiage in the answer which is not asked
for in the question.
Model outputs that match the imprecise ground truth are concerning, and we do find
models where they do. When we test the functionalized, corrected, versions of these
problems the matching dissappears.

These cases illustrate how a functional evaluation separates out various cases,
and provides a more robust evaluation compared to static QAs.

\section{Approach}

Given a benchmark test written as a {\em ``question, answer''} pair, we convert
it to three functions {\em ``problem(inputs), solution(inputs), inputs(rngs,
seed)''}.  These functions are designed to have the property that arriving at
{\em answer} using a well-reasoned derivation from the original {\em question}
is the same that would yield {\em answer' = solution(in)} from {\em question' =
problem(in)} where {\em in = inputs(rngs, seed)} for an arbitrarily chosen {\em
seed}.

Building these functions requires human insight in: a) picking a set of
symbolic inputs that sufficiently generalize both question and answer, b)
encoding any implicit constraints on the inputs in the generator function and
that the generator explores the input space as the seed is varied, c)
translating the text of the specific question and answer, into a respective
functions problem and solution that symbolically derive with the same
steps as would have been followed in the derivation of answer from question.

\paragraph{\em Functional variants of a benchmark} We call a benchmark, e.g.,
MATH, that has its test set converted from static ``question, answer'' form
into functionals as above the {\em functional variant}, noted as MATH(). Each
test in the functional variant cannot be directly used to evaluate, but instead
it needs to be instantiated.

\paragraph{\em Snapshots} We call an instantiation of each of the
functionalized tests using a seed a {\em snapshot} of the benchmark. For our
implementation, we use a hash of the month string, e.g., ``Dec-2023'', combined
with a hash of the original text QA, as the seed. When instantiated, each test
in the snapshot has the same reasoning as the original test but the steps to
the answer will be different. We can run the evaluation scripts unmodified from the
original benchmark over the snapshot.

\paragraph{\em Functional accuracy} We define the functional accuracy as the
accuracy of solving a test in all $k$ snapshots and the static variant.  If a
model arrives at an answer given a question, using proper reasoning, then its
outcome should be identical over the static benchmark as well as each snapshot
of the functional variant. Thus we define the {\em functional accuracy} of the
model against that benchmark as the fraction of tests it gets correct over
$k$-snapshots and the static variant.

\paragraph{\em Reasoning Gap} The reasoning gap is the percent decrease in the
accuracy between the static and functional variants.

\paragraph{\em Example of functionalizing a test}
Figure~\ref{fig:example-text-qa-714} shows an example text QA from the counting and probability subset of the MATH test dataset, that is solved in its
static form by GPT4. While it is not a particularly complicated counting calculation, it does
require more than grade school reasoning of realizing that the problem can be translated into its negated form,
followed by canceling of major terms in the permutation. While it is not impossible that a model would
do step-by-step reasoning of such form if it has seen the logic before, inventing that line of reasoning
from first principles would be surprising. We convert it to a functional test as shown in Figure~\ref{fig:example-fn-qa-714}. Of the three snapshots we take of this, GPT4 solved 1/3 of them.
\begin{figure}
\small
\begin{verbatim}
Text problem:
The letters of the word `SIXTEEN' are randomly arranged.
What is the probability that the two E's are not next to each other?

Text solution:
The best way to do this is to find the probability that the
two E's are next to each other. There are $\\dfrac{7!}{2}$
arrangements of the word SIXTEEN. If we want to find the number
of arrangements such that the E's are next to each other, we find
the number of arrangements of the six-letter word SIXT(EE)N
(where we treat the two E's as a block), which is $6!$.
So the probability that an arrangement of the word SIXTEEN
has the two E's next to each other is
$\\dfrac{6!}{\\frac{7!}{2}} = \\dfrac{2}{7}$.
So the probability that the two E's aren't next to each other is
$1 - \\dfrac{2}{7} = \\boxed{\\dfrac{5}{7}}$.
\end{verbatim}
\caption{\label{fig:example-text-qa-714} Example of a text QA from the MATH test data (Counting and Probability, 714.json). Note that the evaluation suite only compares the correctness of the output text enclosed in ``\bsl\bsl boxed\{..\}''
  in this case the correct answer would be
``\bsl\bsl dfrac\{5\}\{7\}''
  .}
\end{figure}
\begin{figure}
\small
\begin{verbatim}
def problem(word: str) -> str:
    prb = f"The letters of the word `{word}' are randomly arranged."
          f"What is the probability that the two E's are not next to each other?"
    return prb

def solution(word: str) -> str:
    length = len(word)

    # The numerator is always 2 less than the length of the
    # random string - 2 being the length of the string 'EE'
    numerator = length - 2

    # Resulting denominator will always be the length of the string
    denominator = length

    # Simplifying the fraction
    common_factor = math.gcd(numerator, denominator):
    if common_factor > 1:
        numerator //= common_factor
        denominator //= common_factor
    return f"\\dfrac{{{numerator}}}{{{denominator}}}"

def inputs(rngs, seed):
    rngs.set_seed(seed)
    # The string is made up of three parts, pre + 'EE' + post
    # We keep the string 'EE' constant to align with original problem

    # Length of the generated word
    length = rngs.int_between(6, 16)

    # Generate a random index where the two 'E's will be placed
    index = rngs.int_between(0, length - 3)

    # A string containing all uppercase letters except 'E'
    excl_ee = string.ascii_uppercase.replace("E", "")

    # A lambda to build a random word with the given length
    build = lambda length: "".join(rngs.choose_from(excl_ee) for _ in range(length))

    # Construct the string with the three parts
    pre = build(index)
    post = build(length - index - 2)
    word = pre + "EE" + post

    return {"word": word}
\end{verbatim}
\caption{\label{fig:example-fn-qa-714} Functionalization of the text QA from Figure~\ref{fig:example-text-qa-714}}
\end{figure}

Figure~\ref{fig:example-text-qa-1151} shows an example text QA from the prealgebra subset of the MATH test dataset, also solved in its static form by GPT4. Figure~\ref{fig:example-fn-qa-1151} shows its corresponding functionalization. GPT4 reports ``NO SOLUTION'' for each of the snapshots, which it has been instructed to do if it does not know the answer, indicating that its answer to the static version was adding spurious accuracy. While hallucinations are not the core focus of this work, we also test how many times the models report NO SOLUTION, which they rarely do, and this is one of the rare occasions.
\begin{figure}
\small
\begin{verbatim}
Text problem:
Spinner I is divided into four equal sections labeled 2, 3, 4 and 5.
Spinner II is divided into five equal sections labeled 1, 3, 5, 7 and 9.
If each spinner is spun and the resulting numbers are multiplied,
what is the probability that the product is a two-digit even number?
Express your answer as a common fraction.

Text solution:
Let results be denoted by ordered pairs where the first coordinate
corresponds to Spinner I and the second coordinate corresponds to Spinner II.
Since all the section numbers on Spinner II are odd, Spinner I must given an
even number in order for the product to be even. The results
$(2,5)$, $(2,7)$, $(2,9)$, $(4,3)$, $(4,5)$, $(4,7)$, and $(4,9)$
are the ones whose products are two-digit even numbers.
Since there are $5\\times4=20$ equally likely results, the probability of
obtaining an even two-digit product is $\\boxed{\\frac{7}{20}}$.
\end{verbatim}
\caption{\label{fig:example-text-qa-1151} Example of a text QA from the MATH test data (Prealgebra, 1151.json). Note that the evaluation suite only compares the correctness of the output text enclosed in ``\bsl\bsl boxed\{..\}'', in this case the correct answer would be ``\bsl\bsl frac\{7\}\{20\}''.}
\end{figure}

\begin{figure}
\vspace{-2em}
\small
\begin{verbatim}
def problem(spinner1: list, spinner2: list, num_digits: int) -> str:
    # The helper 'csp' formats the list into a comma separated format like
    # in the problem statement '2, 3, 4 and 5', and to_word creates an english word
    return f"Spinner I is divided into four equal sections labeled {csp(spinner1)}. " \
           f"Spinner II is divided into five equal sections labeled {csp(spinner2)}. " \
           f"If each spinner is spun and the resulting numbers are multiplied, " \
           f"what is the probability that the product is a {to_word(num_digits)}-digit " \
           f"even number? Express your answer as a common fraction."

def solution(spinner1: list, spinner2: list, num_digits: int) -> str:
    # The total possible permutations
    total_permutations = len(spinner1) * len(spinner2)

    # The result we need where spinner 1 is even and the product is 'num_digits'
    total_count = sum(
        1
        for i in spinner1
        if i % 2 == 0
        for j in spinner2
        if is_n_digit(i * j, num_digits)
    )

    # Simplifying the fraction
    factor = math.gcd(total_count, total_permutations)
    numerator = total_count // factor
    denominator = total_permutations // factor
    return f"""\\frac{{{numerator}}}{{{denominator}}}"""


def inputs(rngs, seed):
    rngs.set_seed(seed)

    # Generate a random target digit count
    digit_count = rngs.even_between(2, 10)
    one_num_count = digit_count // 2

    lower_limit = 10 ** (one_num_count - 1)
    upper_limit = (10**one_num_count) - 1

    # a, b, c, d are for spinner 1 and e, f, g, h, i are for spinner 2
    # Only numbers generated from more than half of the range of values
    # will multiply to give the double digit of results.
    # So d and f are fixed in having a double digit result.
    # Rest of the products will be random
    a = rngs.int_between(lower_limit, upper_limit)
    b = rngs.int_between(lower_limit, upper_limit)
    c = rngs.natural_int(100000)
    d = rngs.even_between((lower_limit + upper_limit) // 2, upper_limit)
    e = rngs.odd_between(lower_limit, upper_limit)
    f = rngs.odd_between((lower_limit + upper_limit) // 2, upper_limit)
    g = rngs.odd_between(lower_limit, upper_limit)
    h = rngs.odd_between(2, upper_limit)
    i = rngs.odd_between(2, upper_limit)
    spinner1 = [a, b, c, d]
    spinner2 = [e, f, g, h, i]

    # Adding more numbers to the spinners
    for i in range(rngs.int_between(0, 5)):
        spinner1.append(rngs.int_between(lower_limit, upper_limit))

    return {"spinner1": spinner1, "spinner2": spinner2, "num_digits": digit_count}
\end{verbatim}
\caption{\label{fig:example-fn-qa-1151} Functionalization of the text QA from Figure~\ref{fig:example-text-qa-1151}}
\end{figure}

\section{Results and Discussion}

\subsection{Functionalizing MATH to MATH()}
\label{result:functional-math}
We have functionalized 41.2\% (2060/5000) of the MATH benchmark. This subset was chosen based on the
portion of the benchmark that is solvable by a group of SOTA closed and OSS
models, without prompting optimizations (e.g., CoT) and with $pass@1$. This
choice makes the evaluation complete for the current set of models, because by
definition tests that fail the static version do not count towards overall
functional accuracy. The next release of the benchmark will provide 100\%
coverage, and also permit interpretable results under model-specific prompting
optimization, and options for testing against $pass@k$ or $maj@j$ for $k,j>1$.

\subsection{Reasoning gap for major models}
\label{result:gap}

\paragraph{Models evaluated} We ran evaluations over SOTA reasoning models: The closed models accessed through APIs were OpenAI's GPT3.5 and GPT4~\cite{gpt4}, Anthropic's Claude 2.1~\cite{claude2.1}, and Mistral's Mixtral Medium, 7x8B MoE and 7x8B MoE Instruct~\cite{mixtral}. The OSS models included were LLaMA 2 70B~\cite{llama-2}, WizardCoder Python 34B~\cite{wizardcoder-34B}, Yi 34B and Yi Chat 34B~\cite{yi}, StripedHypena Nous and Hessian 7B~\cite{striped-hyena}, Qwen 7B~\cite{qwen}. We chose this subset of models to sample a diversity of sizes, architectures, training data and recipes, and instruction and preference tuning.

\begin{figure*}
  \begin{tabular}{c}
    \includegraphics[width=0.80\textwidth]{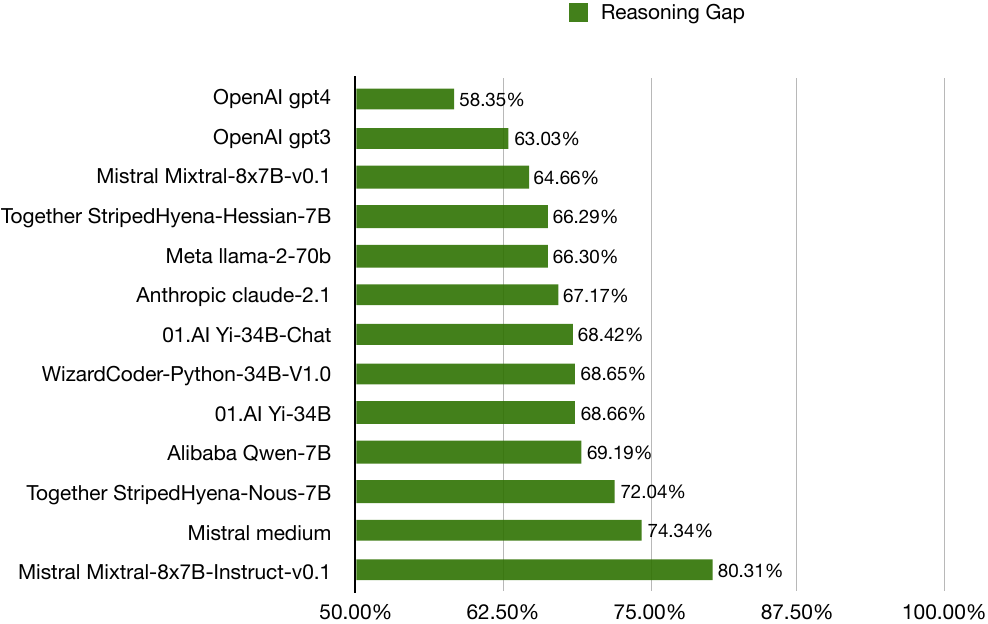}\\
    (a)\\\\
    \includegraphics[width=0.70\textwidth]{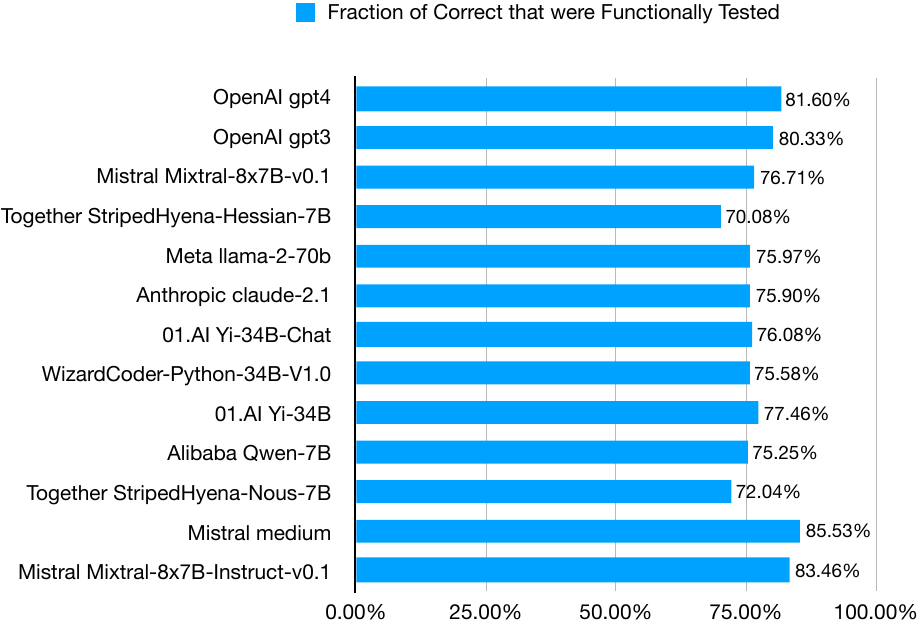}\\
    (b)\\\\
    \includegraphics[width=0.85\textwidth]{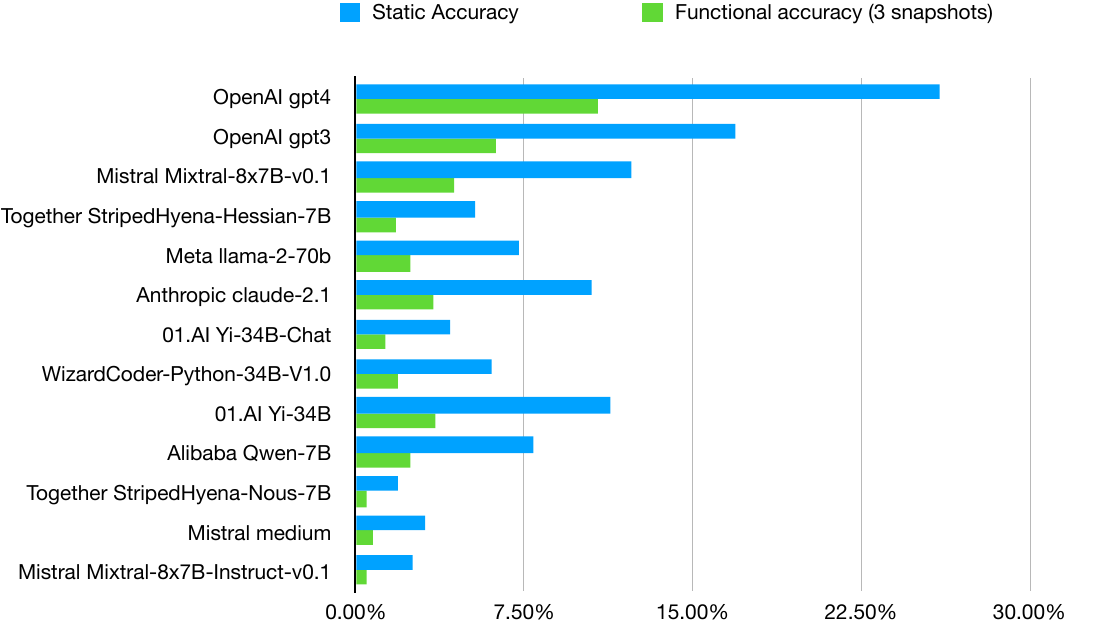}\\
    (c)
  \end{tabular}
  \caption{\label{fig:accuracy-gap} (a) Reasoning gap (note: x-axis starts at 50\%), (b) Coverage: fraction of static QA that
  the model solves correctly that are tested functionally, (c) Static and functional accuracies.}
\end{figure*}

\paragraph{Gap = 58-80\% for SOTA models} Evaluated over the Q1'24 snapshot (consisting of Oct-2023, Nov-2023, Dec-2023), we find that the models have a reasoning gap varying between 58.35\% and 80.31\%, as shown in Figure~\ref{fig:accuracy-gap}(a). Figure~\ref{fig:accuracy-gap}(b) shows the percentage of problems the model solved correctly that were evaluated using functional snapshots. While we attempted to functionalize every problem, some were not convertible, either because the problem was already as general as possible, or it was so specific that it did not permit any parameterization. Figure~\ref{fig:accuracy-gap}(c) shows the individual static accuracy vs functional accuracy.

\begin{figure*}
  \begin{tabular}{c}
    \includegraphics[width=\textwidth]{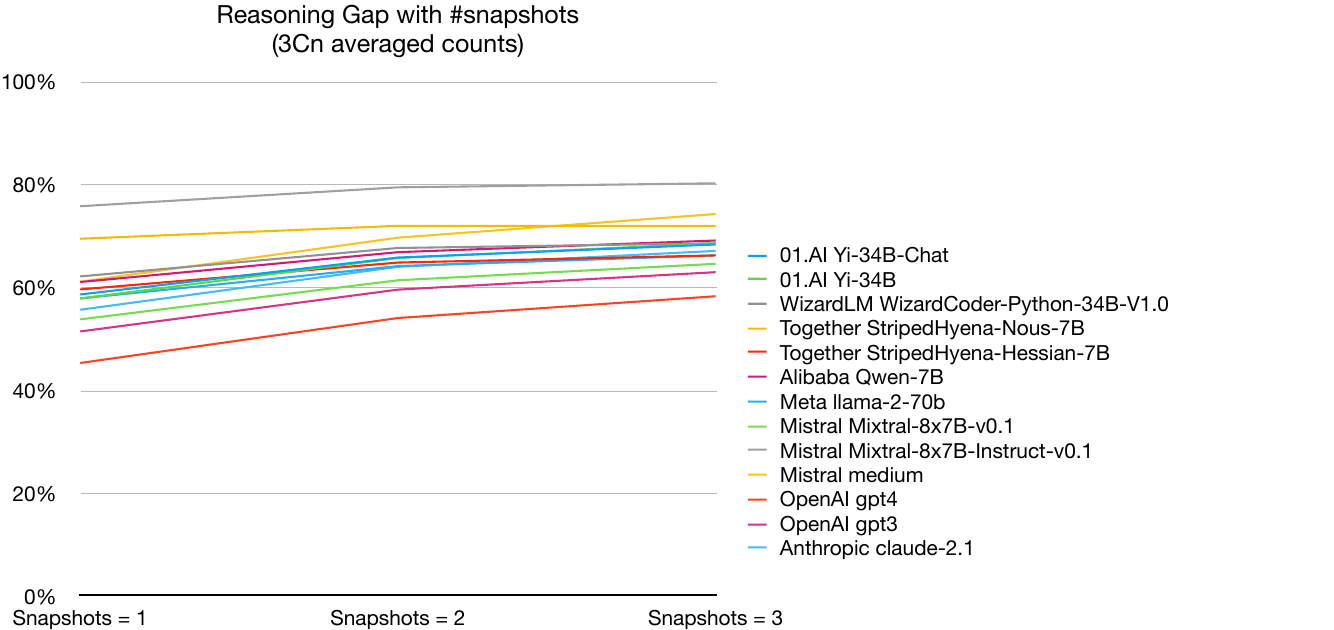}\\
    (a)\\\\
    \includegraphics[width=0.6\textwidth]{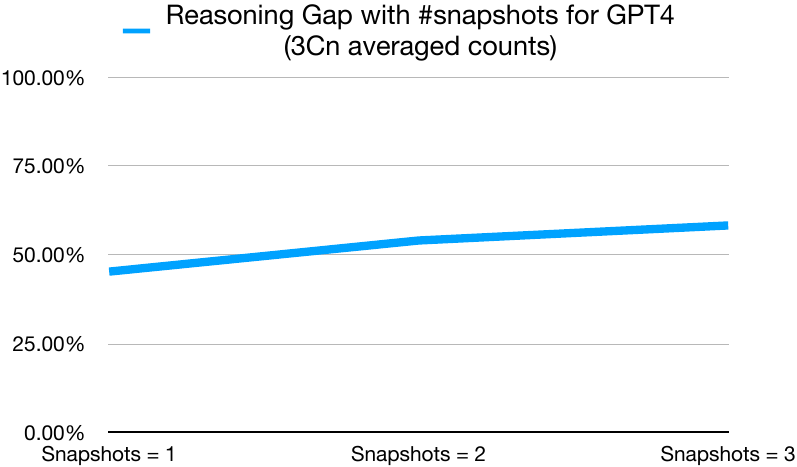}\\
    (b)
  \end{tabular}
  \caption{\label{fig:gap-stabilizes-k-3} Gap starts at close to final value with a single snapshot, increases slightly with 2 snapshots, and essentially stabilizes at 3 snapshots.}
\end{figure*}

\paragraph{Gap stabilization after three snapshots}
Figure~\ref{fig:gap-stabilizes-k-3} shows our analysis of gaps with three
snapshots, two snapshots (three subsets chosen from Oct, Nov, Dec), and one snapshot
(Oct, Nov, Dec taken individually).  For two and one snapshot subsets of which
there are three each, we take the mean as the representative, since we observed
only minor variation in the accuracy.  Raw data for all subsets is available in
the associated repository for further examination.
Figure~\ref{fig:gap-stabilizes-k-3}(a) shows the gaps for all models, and
Figure~\ref{fig:gap-stabilizes-k-3}(b) shows GPT4's gap separately.  We find
that three snapshots suffice to stabilize the reasoning gap. Including further
snapshots does not materially alter the gap.  We take this to mean that tests
that are solved by the models consistently with $k=3$ are solved by reasoning
about the problem. The likelihood of accidentally solving three separate
snapshots, including the static variant, is miniscule.

\paragraph{Gap across difficulty levels in MATH()}
\begin{figure*}
  \begin{tabular}{cc}
    \includegraphics[width=0.5\textwidth]{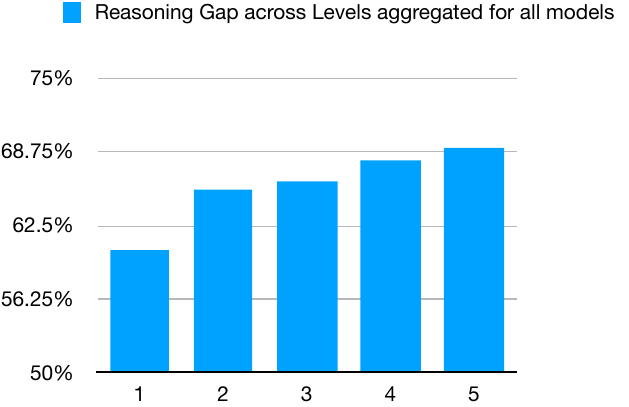} &
    \includegraphics[width=0.5\textwidth]{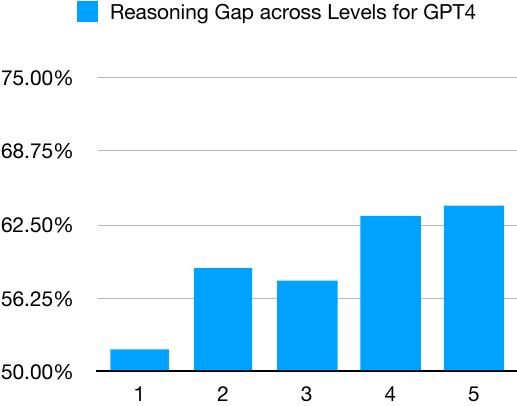}\\
    (a) & (b)
  \end{tabular}
  \caption{\label{fig:gap-levels} Gap across levels, aggregated across all models and for GPT4 separately.}
\end{figure*}
MATH has problems categorized into five levels.  Figure~\ref{fig:gap-levels}
shows the gap against these levels. The levels in the test dataset roughly
correspond to the difficulty of the problems, although our subjective
assessment indicates that levels 1-3 are hard to discern as distinct from each
other.  We find that the reasoning gap expectedly increases with difficulty
levels, suggesting that the models are capable of simpler reasoning at the
lower difficulty levels and solve harder problems with more memorization
(Figure~\ref{fig:gap-levels}(a)).  The trend is similar for the top model
(GPT4), except its gap is lower for each individual level
(Figure~\ref{fig:gap-levels}(b)).

\paragraph{Gap across subject levels in MATH()}
\begin{figure*}
  \begin{tabular}{cc}
    \includegraphics[width=0.5\textwidth]{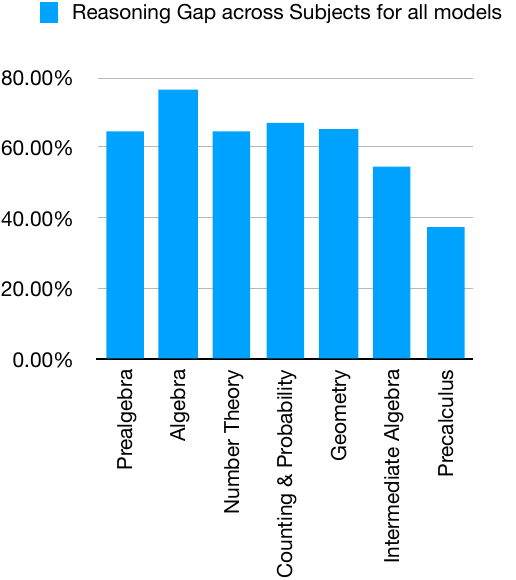} &
    \includegraphics[width=0.5\textwidth]{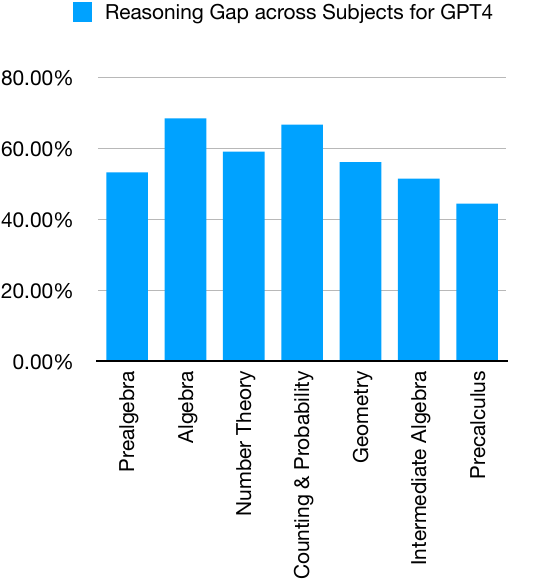}\\
    (a) & (b)
  \end{tabular}
  \caption{\label{fig:gap-subjects} Gap across subjects, aggregated across all models and for GPT4 separately.}
\end{figure*}
Figure~\ref{fig:gap-subjects} shows the gap against subjects. Precalculus, prealgebra, and intermediate algebra
are the easiest problems in the dataset and the reasoning gap is expectedly lowest for those, indicating
again that the models are solving problems under those subjects with proper reasoning (Figure~\ref{fig:gap-subjects}(a)). Similar trends hold across
the aggregated set of models and in the top model (GPT4), although GPT4's gap is lower across all subjects (Figure~\ref{fig:gap-subjects}(b)).

\subsection{Analysis of problems solved across all snapshots, i.e., with proper reasoning}
\label{result:fn-solved}

\paragraph{Problems solved by the top model (GPT4)}
GPT4 solves 1299 static QA problems (accuracy = 25.98\%), and its functional accuracy with 3-snapshots is 10.82\% (541/5000), giving it the lowest reasoning gap of 58.35\% amongst the models tested. Of these:
\begin{itemize}
  \item 239 were ungeneralizable, and their functional versions are static. This means that 4.78\% of its functional accuracy can be attributed to being static.
  \item 302 had their functional snapshots solved, giving it a 6.04\% accuracy on problems that indeed were different across snapshots.
\end{itemize}
We manually categorized each of the 302 that were solved based on a) hardness as indicated by levels and subjects, b) our {\em subjective} surprise (low, medium, high) that they were solved based on the individual steps and process required to solve them, c) our {\em subjective} tag of the type of reasoning needed to solve.
\begin{figure*}
  \begin{tabular}{cc}

    \includegraphics[width=0.4\textwidth]{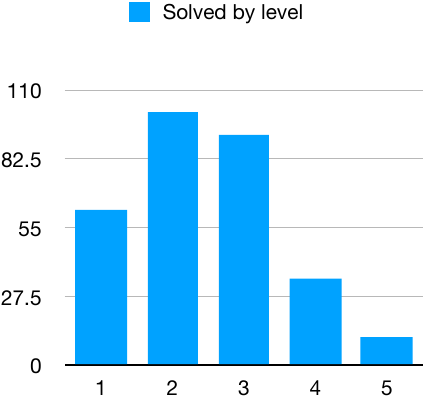} &
    \includegraphics[width=0.4\textwidth]{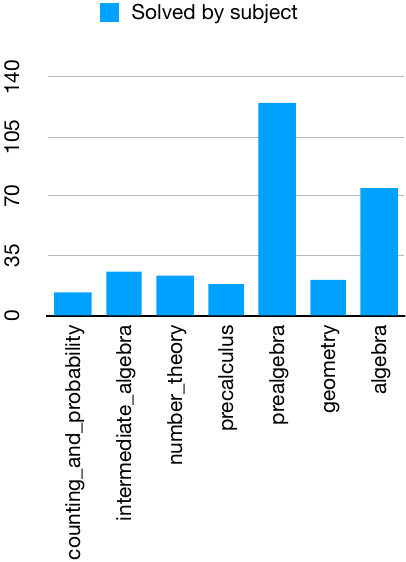} \\
    (a) & (b) \\\\

    \includegraphics[width=0.4\textwidth]{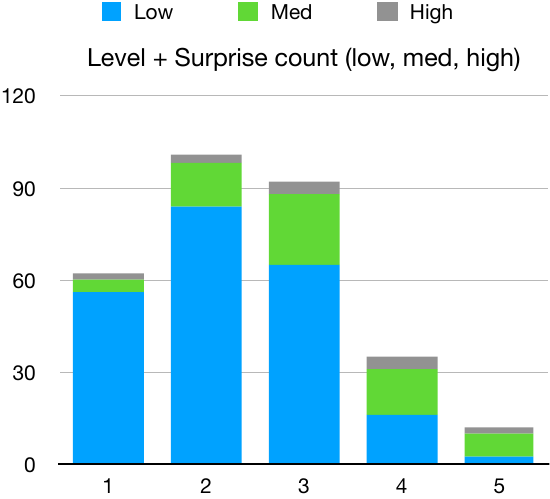} &
    \includegraphics[width=0.5\textwidth]{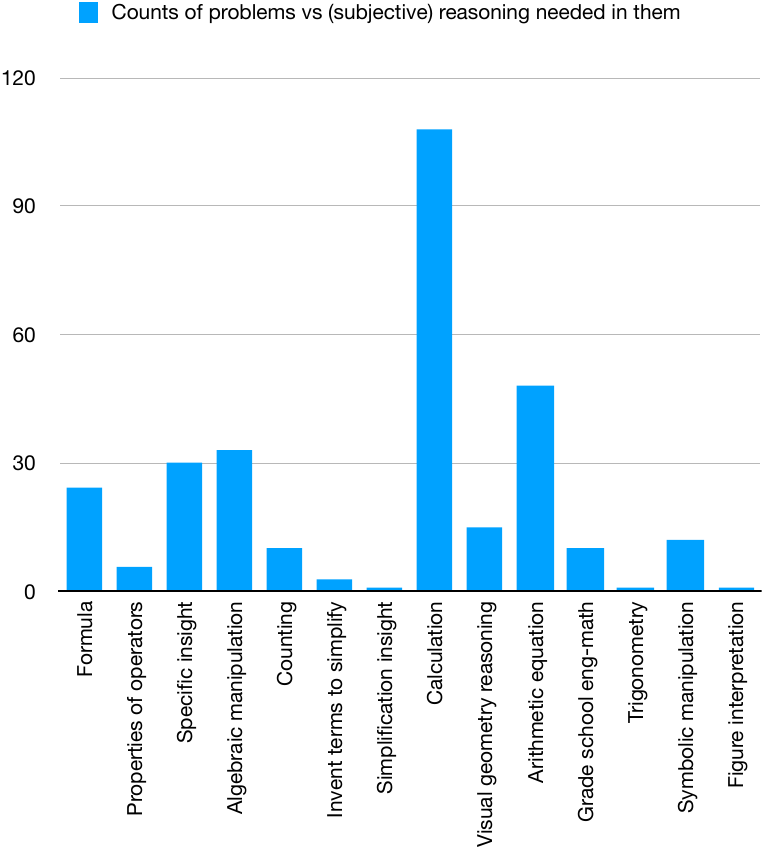} \\
    (c) & (d)

  \end{tabular}
  \caption{\label{fig:gpt4-fn-solved} 302 problems were solved by GPT4 across all snapshots.}
\end{figure*}

\paragraph{Solved hardness, by levels:} We find an expected dropoff across
levels (Figure~\ref{fig:gpt4-fn-solved}(a)), except for level 1. Human
assessment from the team of mathematicians and computer scientists doing the
functionalization, we find little difference between levels 1 and 2, and if we
aggregate them then we find a monotonic drop in numbers solved.

\paragraph{Solved hardness, by subjects:} For subjects
(Figure~\ref{fig:gpt4-fn-solved}(b)), prealgebra and algebra are outliers in
being easier to solve. Most of the problems here are of the ``grade school
math'' style problems where the task is to interpret the english description
into a simple arithmetic equation. Their high solved counts here reinforce the
commonly held perception that SOTA models are good at solving those problems.

\paragraph{224 low-, 63 medium-, 15 high-surprise problems solved:} We find 2-4
problems with high surprise (15 in total) distributed across all levels
(Figure~\ref{fig:gpt4-fn-solved}(c)). It might be instructive to the community
with access to the model and training dataset to understand the model's ability
in solving these. The distribution of medium surprise problems follows an
expected normal distribution.

\paragraph{14 solving strategies:} We identify 14 {\em subjective} subtypes
(Figure~\ref{fig:gpt4-fn-solved}(d)) of solving strategies that would be needed
for the 302 problems. Of these the extremes are worth noting. The highest
count is for {\em simple calculations} by a significant margin, which aligns
with capabilities, especially, if access to a calculator is in-built. The
lowest counts are for {\em figure interpretation} (1 instance), {\em
trigonometry} (1 instance), {\em simplification insight} (1 instance), and {\em
invent terms to simplify} (3 instances). All these are worthy of individual
investigation as they indicate surprising capabilities.
Appendix~\ref{appendix:gpt4-fn-solved-instances} lists a) two illustrative
examples for each of the 14 reasoning subtypes, and b) the 15 specific {\em
high-surprise} problems solved.

\begin{figure*}
  \begin{tabular}{cc}

    \includegraphics[width=0.4\textwidth]{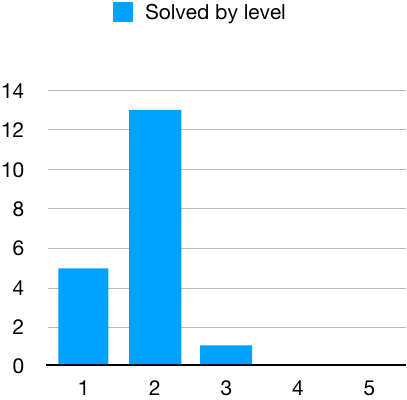} &
    \includegraphics[width=0.4\textwidth]{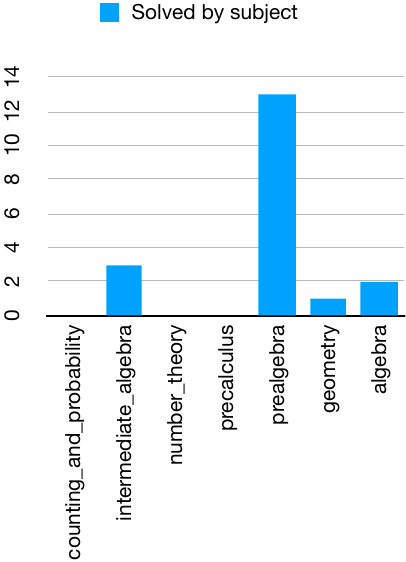} \\
    (a) & (b) \\\\

    \includegraphics[width=0.4\textwidth]{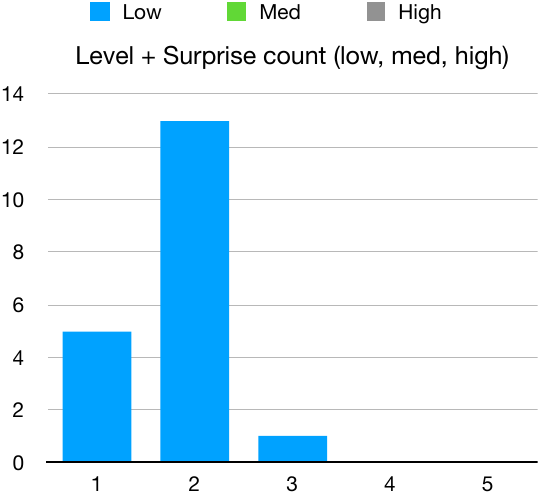} &
    \includegraphics[width=0.4\textwidth]{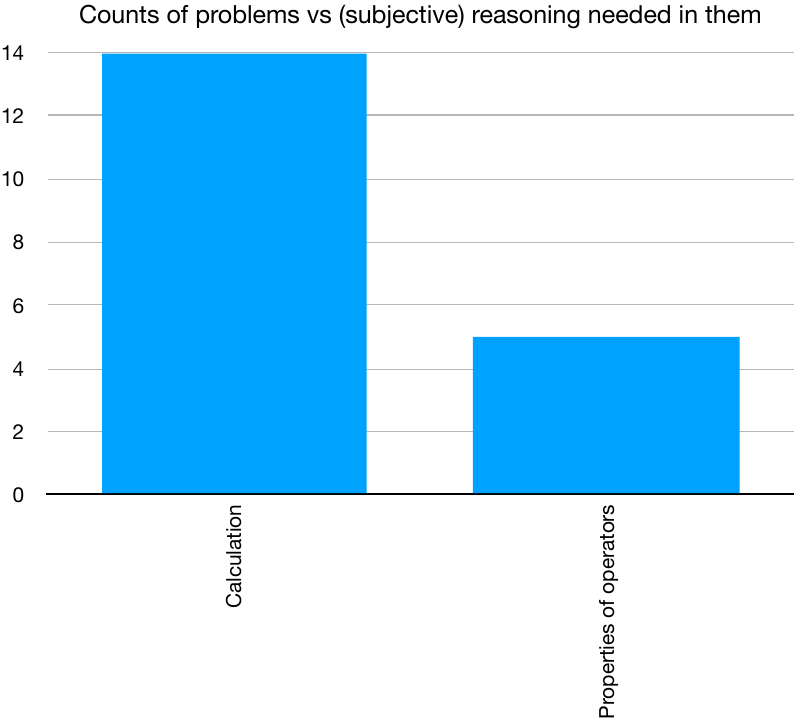} \\
    (c) & (d) \\

  \end{tabular}
  \caption{\label{fig:oss-fn-solved-analysis} Problems solved by the majority ($\ge 5/10$) of OSS models.}
\end{figure*}

\paragraph{Problems solved by a majority of the OSS models}
The accuracy of OSS models varies between 1.86\%-12.28\% and it drops to
0.52\%-4.34\% for the functional snapshots. Only 19 problems are consistently
solved across snapshots by the majority ($\ge 5/10$) of models.

\paragraph{Solved hardness, by levels:} As argued earlier, levels 1 and 2 are
not substantially different, and we find that OSS models essentially solve
these levels, except for one outlier in level 3
(Figure~\ref{fig:oss-fn-solved-analysis}(a)).

\paragraph{Solved hardness, by subjects:} OSS models solutions are limited to
the easier subjects of prealgebra, algebra, intermediate algebra, and a single
solution in geometry (Figure~\ref{fig:oss-fn-solved-analysis}(b)).

\paragraph{19 low-surprise problems solved:} There are no medium or high surprise
problems consistently solved by the majority of the models
(Figure~\ref{fig:oss-fn-solved-analysis}(c)).

\paragraph{2 reasoning subtypes:} The problems solved consistently were either
simple calculations, or involving use of simple properties of operators
(Figure~\ref{fig:oss-fn-solved-analysis}(d)).
Appendix~\ref{appendix:oss-fn-solved-instances} list a) 14 problems solved
which were simple calculations, b) 5 problems which required properties of
operators.

\section{Threats to the validity of results}
\begin{itemize}
  \item {\em More sophisticated prompting will likely change precise reasoning
    gap:} Model-specific prompting (e.g., Claude2.1 retrieval
    prompting~\cite{claude-retrieval-prompting}) or model-agnostic methods such
    as chain-of-thought~\cite{chain-of-thought},
    tree-of-thought~\cite{tree-of-thought}, chain-of-code~\cite{chain-of-code},
    chain-of-thought-decoding~\cite{cot-decoding},
    tipping/threats/persuasion prompting to elicit more accurate output, are
    likely to reduce the reasoning gap. In fact, even if accuracy remains the
    same, the reduced reasoning gap may quantitatively validate the utility of
    these improvements.

  \item {\em Tool usage:} For closed weights models accessed through APIs, it is
    unclear whether tool usage is part of the inference pipeline. We expect the
    use of tools, e.g., calculators or theorem provers, to reduce the reasoning
    gap. In future work, we will evaluate the change in gap with or without
    tool use with open weight models.

  \item {\em Default chain of thought, i.e., unable to format according to
    few-shot format:} Some models seem to not follow instructions, or learn in-context
    with few shot examples, as others. In particular, we find Anthropic's Claude 2.1 and
    Mistral models tend to default output chain-of-thought steps, which does not
    work very well with the evaluation harness.

  \item {\em Scenarios where functional variant is not general enough}: When we
    could not generalize a test, we marked it as ``is static''.  This could
    happen at two extremes: i) when the problem is already in symbolic form or,
    ii) only permits a single formulation. These labelled cases are the reason
    for less than 100\% coverage in Section~\ref{result:gap}. Aside from these
    cases, there might be scenarios where we functionalize but the generalization
    is simple enough that a pattern match against the original static QA would
    permit solving the functional snapshots. These cases should be rare.

  \item {\em Prompt injection in closed source models or non-determinism in
    closed source APIs:} We do not have access to the additional system prompts being
    added by the closed source APIs for safety, which might cause reduced accuracy.

  \item {\em Big numbers:} On rare occasions the random number functions might
    instantiate an excessively large number, e.g., with 10+ digits, and it would be unreasonable for the model
    to compute that even if the reasoning used was correct.

  \item {\em Output matching:} The evaluation code attempts to be liberal in equivalence
    checking for outputs, but has noticeable limitations. Cases we observe, and
    might be hard automatically check for, but would be judged as correct by a
    human evaluator include: a) \verb|C(21,7)| and \verb|9 choose 2| instead of
    their evaluated counterparts, b) expanded output expressions such as
    \verb|5! = 5 x 4 x 3 x 2 x 1| and \verb|33(22-12+1)=33*11=363|, c)
    extra formatting characters in output such as LaTeX math delimiters
    \verb|$\\frac{1}{16}$|,
    d) extra english verbiage such as
    \verb|g(x) = 3 - 2f(x)|'' (for ground truth ``\verb|3 - 2f(x)|'') and\\
    ``\verb|The remainder when $2^8$ is divided by 5 is 1.|''.

\end{itemize}

\section{Related Work}

\paragraph{Reasoning benchmarks}
Many popular benchmarks exist for testing reasoning, focussing on subdomains
such
abstract reasoning (ARC~\cite{cholet-arc}),
mathematical (MATH~\cite{math} and GSM8K~\cite{gsm8k}),
code (HumanEval~\cite{humaneval}, MBPP~\cite{mbpp}, and Natural2Code~\cite{gemini}),
commonsense and world knowledge
(HellaSwag~\cite{hellaswag},
Winogrande~\cite{winogrande}, PIQA~\cite{piqa}, SIQA~\cite{siqa},
OpenbookQA~\cite{openbookqa}, ARC-Easy/Challenge~\cite{arc-easy-challenge},
CommonsenseQA~\cite{commonsenseQA},
NaturalQuestions~\cite{naturalquestions}, TriviaQA~\cite{triviaQA}),
logical (LSAT~\cite{lsat}),
or legal (LegalSupport~\cite{helm}).
Aggregated benchmarks collect many of these domains and attempt to give
a comprehensive view of a generalized capabilities for foundational models
(MMLU~\cite{mmlu},
HELM~\cite{helm}, BBH~\cite{big-bench-hard}, and AGI Eval~\cite{agi-eval}).
The aggregated benchmarks cover too much ground and might not be the best
metrics for reasoning.
There are concerns of contamination and overoptimization towards popular
benchmarks leaving a gap between evaluated capabilities, and experienced
capabilities over real-world tasks.

To alleviate that concern, there has been a push towards building new, and
usually more difficult, text QA benchmarks for science and math (e.g.,
GPQA~\cite{gpqa}, ARB~\cite{arb}, JEE~\cite{jee}, CLRS~\cite{clrs}, ProofNet~\cite{proofnet}), agentic and world model reasoning (e.g.,
GAIA~\cite{gaia}, WorldSense~\cite{worldsense}) and code (e.g.,
CRUXEval~\cite{cruxeval}, SWE-Bench~\cite{swe-bench}, xCodeEval~\cite{xcodeeval}).
A recent survey~\cite{survey-reasoning} provides a comprehensive list of
reasoning tasks by categories.
We worry that without a
systemic move away from text QAs, such benchmarks are prone to eventual
leakage, and overestimation of solving capabilities of existing models.

Alternative benchmarking techniques are being proposed, e.g., through
uncertainty quantification~\cite{uncertainity-quant}, ROSCOE step-by-step scoring~\cite{roscoe},
counterfactual facts~\cite{counterfactual}.

We would suggest using our functionalization or the alternative benchmarking
techniques, applied over existing popular or new text benchmarks as the way
forward.

\paragraph{Techniques to improve reasoning in language models}
There is a vast body of ongoing work on improving reasoning capabilities of language models. Especially modifications to training data, or fine tuning recipes. We do not intend cover that literature. Instead, we mention techniques that might impact the reasoning gap during inference, assuming a given base model.
\begin{itemize}
  \item Specialized prompting to elicit explicit reasoning:
      Techniques such as chain of thought~\cite{chain-of-thought},
      tree of thought~\cite{tree-of-thought},
      chain of code~\cite{chain-of-code},
      chain of hindsight~\cite{chain-of-hindsight},
      program of thought~\cite{program-of-thought},
      algorithmic skill prompting~\cite{alg-prompting},
      progressive hint prompting~\cite{progressivehint} give models more
      scratch space to make
      the reasoning logic explicit, and arguably decompose the argument into
      simpler more tractable steps. They have been shown to improve accuracy,
      and we would conjecture they would reduce the gap. In future work we will
      explore how they change the gap.

  \item Augmentation: Delegating the last mile of the inference to a
    tool~\cite{tora, talm} (calculator, web search~\cite{react}, theorem prover,
    symbolic solver~\cite{symbolicai}, planner~\cite{llmp}, interpreter~\cite{pal}) is bound to help
    reduce the reasoning gap if the model
    knows how to decompose the task into appropriate symbolic form.

  \item Self-inspection or related post-processing to improve generalization:
    Various techniques have been proposed for using the core feedforward
    network as a unit module around with an inference pipeline can be built.
    These include using indirect reasoning~\cite{indirect-reasoner},
    self-critique~\cite{self-critique}, divide-and-conquer~\cite{dnc},
    sampled math or code prompting~\cite{math-prompter},
    planner guided decoding~\cite{code-planner-decoding},
    self-consistency~\cite{self-consistency},
    recursive code template improver~\cite{recursive-code-improver}
    logic guide-driven inference~\cite{logic-guide},
    self-debug~\cite{self-debug},
    least-to-most prompting~\cite{least-to-most},
    self-discover to compose reasoning structures~\cite{self-discover}.
    It is
    an open problem how much the accuracy improvements using these techniques
    translate to lowered reasoning gaps.

\end{itemize}

\paragraph{Understanding the bounds of reasoning, generalization, and
memorization in large language models}
Evaluations are the guides against which we build. They test the end artifact.
If we need improved reasoning in these end artifacts, using reasoning-specific
training data has shown to help---e.g., ORCA~\cite{orca,orca2},
large models as reasoning teachers~\cite{large-reasoning-teachers},
textbook quality data~\cite{textbooks}, and code data~\cite{code-training-helps}.
Additionally, reasoning-specific fine-tuning
algorithms help improve performance after pre-training---e.g.,
process supervision~\cite{process-supervision,math-shepherd},
self-distilling context~\cite{self-distilling-context},
feedback training for math using oracle teachers~\cite{rest-self-training-oracle, rest-multistep-self-training},
self-play fine-tuning for weak to strong learning~\cite{self-play-for-weak-to-strong},
refusal-aware fine tuning to reduce hallucinations~\cite{refusal-aware},
improving reasoning by removing higher order components~\cite{laser-remove-higher-order-weight-components},
and self-instruct~\cite{self-instruct}.

At the other end of the spectrum are {\em investigative analyses}
that improve our understanding.
We now know it is possible to extract training
data~\cite{extract-training-data} so some memorization is inevitable, and that
the notion of emergent properties might have been a result of choice of
metrics~\cite{emergent-mirage}. Task contamination is an issue we need to
be careful of~\cite{task-contamination}.  Some encouraging results show evidence
of multi-hop reasoning~\cite{multi-hop-reasoning} while others demonstrating
difficulty of such reasoning~\cite{reasoning-multiple-facts}.  Evidence exists
for transformer models identifying training data patterns very robustly, while
performance degrading outside of distribution~\cite{ood-hard}. Reasoning might
emerge from path-aggregation in the training
data~\cite{path-aggregation-leads-to-reasoning}, and that asking for lengthier
answers may improve reasoning~\cite{length-improves}.  Models are unlikely to
know when they are violating formal rules and it is unclear whether they can
self-correct~\cite{cant-find-errors-but-correct,cannot-self-correct,dont-know-wrong},
but with specific fine-tuning they might self-correct against harmful
text~\cite{moral-self-correction}, and that training on generated data might
not be the best approach to preserve reasoning about outlier
cases~\cite{gen-data-makes-models-forget}.

Given such evolving understanding, a proper scientific approach should
innovate on evaluations based on the most recent understanding,
so that progress stays stays aligned with real-world experience.

\section{Individual contributors and acknowledgments}
Annarose, Shashank, Anto, Ajay, Adwaith, Alan, and Stevin did the major work of converting each
MATH test into code, and Sooraj oversaw their work.
Saurabh conceptualized the functionalization and reasoning gap framework,
implemented the evaluation code, and wrote the paper.
Feedback from initial reviewers Kelly Iknayan, Ashish Agarwal, Prashast
Srivastava, Henele Adams, and Soham Mazumdar, helped improve the manuscript.

\section{Future work}
This is part of an ongoing effort to build reasoning metrics that are robust against contamination, and a more accurate representation of model capabilities. In subsequent work, we will functionalize the entire MATH test suite of 5000 problems, and the 1000 GSM8K test problems. We will also use a similar strategy to functionalize the code benchmarks HumanEval, and MBPP.

Additionally, once the benchmarks are 100\% functionalized, we will do studies on the effect of prompting and augmentation strategies (e.g., chain-of-thought, chain-of-code, tree-of-thought, tool usage) on the reasoning gap.

\section{Conclusion}
There is a disconnect between the high benchmark scores in reasoning and the observed below-average reasoning of state-of-the-art models. We propose the {\em reasoning gap} metric to quantify this difference, which motivates the open problem of building {\em gap-0} models. Reasoning gaps can be evaluated using the framework of functionalized benchmarks, which are presented as a long term solution to evaluating reasoning on tasks that are known to be within the scope of current models, while presenting them versions they have not seen before. We have functionalized the relevant portion of the popular MATH benchmark, called MATH(), and in future work will extend this to other gold-standard reasoning benchmarks. We are releasing the evaluation code, and the Q1'24 snapshot of MATH(), and will continue to release new snapshots of functionalized benchmarks every quarter.

\bibliographystyle{plain}
\bibliography{refs}

\section{Appendix: Examples functionally solved by GPT4}
\label{appendix:gpt4-fn-solved-instances}

In this section, we give examples of problems solved by GPT4 across all snapshots, which indicates that the
inference pipeline has the ability to solve the problem in its generalized form.
In the first subsection, we do a subjective categorization of types of reasoning and give two examples
of problems in each category. In the second subsection we cover problems that are solved, but
seem counterintuitive to be solvable with an LLM.

\subsection{A subjective categorization of 14 types of problems GPT4 is capable of handing}

Below we list 14 categories and two examples within each category.
\begin{enumerate}
  \item Formula (24 solved)
    \begin{itemize}
      \item intermediate\_algebra/1231.json: Evaluating the magnitude of a complex number
\begin{verbatim}
Find the magnitude of the complex number $5-12i$.
\end{verbatim}

    \item geometry/514.json: Relating the diameter of a cylinder to surface of flat side
\begin{verbatim}
What is the radius, in inches, of a right circular cylinder
if the lateral surface area is $24\\ pi$ square inches
and the volume is $24\\ pi$ cubic inches?
\end{verbatim}
    \end{itemize}

  \item Properties of operators (6 solved)
    \begin{itemize}
      \item intermediate\_algebra/149.json: Radical conjugate, product gets ($a^2-b^2$); substitution and evaluation
\begin{verbatim}
Compute the product of the number $5+\\sqrt{3}$ and
its radical conjugate.
\end{verbatim}
      \item counting\_and\_probability/64.json: Expected value of independent trials is sum of expectations
\begin{verbatim}
Two fair eight-sided dice have their faces numbered from
1 to 8. What is the expected value of the sum of the
rolls of both dice?
\end{verbatim}
    \end{itemize}

  \item Specific insight (30 solved)
    \begin{itemize}
      \item intermediate\_algebra/1870.json: Terms accumulate sum over all previous values, implies geometric series
\begin{verbatim}
A sequence of positive real numbers $\\{a_1, a_2, a_3, \\dots\\}$
has the property that for $i \\ge 2,$ each $a_i$ is equal to the
sum of all the previous terms.  If $a_{19} = 99,$ then what is $a_{20}$?
\end{verbatim}
      \item algebra/2588.json: Degree of the sum or difference of two polynomials is the greater of the components, if the greater is strictly greater
\begin{verbatim}
If $f(x)$ is a polynomial of degree 4, and $g(x)$ is a polynomial
of degree 2, then what is the degree of polynomial $f(x) - g(x)$?
\end{verbatim}
    \end{itemize}

  \item Algebraic manipulation (33 solved)
    \begin{itemize}
      \item algebra/1561.json: Simplifying an expression and eliminate variables
\begin{verbatim}
There are real numbers $A$ and $B$ such that\n
\\[\\frac{5x-16}{x^2-7x+10}=\\frac{A}{x-2}+\\frac{B}{x-5}.\\]
Find $A+B$.
\end{verbatim}
      \item number\_theory/100.json: Gcd-lcm product relationship and algebraic manipulation
\begin{verbatim}
Find $n$ if $\\gcd(n,40) = 10$ and
$\\mathop{\\text{lcm}}[n,40] = 280$
\end{verbatim}
    \end{itemize}

  \item Counting (10 solved)
    \begin{itemize}
      \item prealgebra/439.json: Counting the length of a list of numbers
\begin{verbatim}
How many numbers are in the list $25, 26, 27, \\ldots, 99, 100 ?$
\end{verbatim}
      \item prealgebra/1703.json: grade-school math; counting possibilities and computing probability
\begin{verbatim}
Two fair 6-sided dice are rolled.  What is the probability
\"doubles\" are rolled (i.e., that the two dice show the same number)?
\end{verbatim}
    \end{itemize}

  \item Invent terms to simplify (3 solved)
    \begin{itemize}
      \item intermediate\_algebra/898.json: Adding a magic term and finding it leads to bound on term whose max requested; max at equality
\begin{verbatim}
If $x$ and $y$ are real, and $x^2 + y^2 = 1,$
compute the maximum value of $(x + y)^2.$
\end{verbatim}
      \item intermediate\_algebra/838.json: Noticing that subtracting two specific given cubic polynomials reduces their degrees to a square with +a, -a roots, Vieta’s formula then used to find root of cubic
\begin{verbatim}
The equations $x^3 + 5x^2 + px + q = 0$ and
$x^3 + 7x^2 + px + r = 0$ have two roots in common.
If the third root of each equation is
represented by $x_1$ and $x_2$ respectively,
compute the ordered pair $(x_1,x_2).$
\end{verbatim}
    \end{itemize}

  \item Simplification insight (1 solved)
    \begin{itemize}
      \item intermediate\_algebra/709.json: Product of fractions where numerator and denominator cancel has a simple outcome
\begin{verbatim}
What is the value of $\\frac{2}{3} \\times \\frac{3}{4}
\\times \\frac{4}{5} \\times \\dotsm \\times \\frac{8}{9}$?
Express your answer as a common fraction.
\end{verbatim}
    \end{itemize}

  \item Calculation (108 solved)
    \begin{itemize}
      \item prealgebra/1364.json: 5\% x 10\% x 1200
\begin{verbatim}
What number is $5\\%$ of $10\\%$ of 1200?
\end{verbatim}
      \item prealgebra/1552.json: Definition of complement of an angle
\begin{verbatim}
The complement of angle $M$ is 10 degrees.
What is the measure, in degrees, of angle $M$?
\end{verbatim}
    \end{itemize}

  \item Visual geometry reasoning (15 solved)
    \begin{itemize}
      \item intermediate\_algebra/470.json: Quadratic equation; limited to one half of the equation makes it invertible
\begin{verbatim}
The function $f(x) = -3x^2 + 36x - 7,$ defined for all real numbers,
does not have an inverse.  However, when its domain is restricted to
$x \\in [c,\\infty),$ the function does become invertible, for
certain values of $c.$  What is the smallest such value of $c$?
\end{verbatim}
      \item prealgebra/994.json: grade-school math; visual drawing of placements on circle
\begin{verbatim}
Bobby stands 10 feet from Sam, and Eric stands 8 feet from Bobby.
What is the least number of feet that Eric could be from Sam?
\end{verbatim}
    \end{itemize}

  \item Arithmetic equation (48 solved)
    \begin{itemize}
      \item prealgebra/1000.json: $5x < -32$, smallest x
\begin{verbatim}
What is the smallest multiple of $5$ which is greater than $-32$?
\end{verbatim}
      \item prealgebra/1262.json: Express an English sentence as an arithmetic equation and solve
\begin{verbatim}
John meant to divide a number by $2$, but he was careless and
subtracted $2$ from the number instead.  He got an answer of $22$.
What would his answer have been had he actually divided by $2$
instead of subtracting $2$?
\end{verbatim}
    \end{itemize}

  \item Grade school english math (10 solved)
    \begin{itemize}
      \item prealgebra/1922.json: Fractions
\begin{verbatim}
Two-thirds of the students at Baker Middle School take music.
There are 834 students who take music.
How many students are there at Baker Middle School?
\end{verbatim}
      \item prealgebra/75.json: Converting simple logic into an arithmetic expression and solving
\begin{verbatim}
Cheldelin Middle School has 12 doors to enter or leave the building.
In how many ways is it possible to enter the building by one door
and leave the building by a different door?
\end{verbatim}
    \end{itemize}

  \item Trigonometry (1 solved)
    \begin{itemize}
      \item precalculus/1185.json: (a-b)(a+b) property and algebraic simplification with relationship between sec/tan/cost/sin; our functionalization is naive (factor added)
\begin{verbatim}
Let $x$ be a real number such that $\\sec x - \\tan x = 2.$
Find $\\sec x + \\tan x.$
\end{verbatim}
    \end{itemize}

  \item Symbolic manipulation (12 solved)
    \begin{itemize}
      \item algebra/2126.json: Simplify expression
\begin{verbatim}
Suppose $x$ is a solution to $x^2 + 1 = 7x$.
What is the sum of $x$ and its reciprocal?
\end{verbatim}
      \item algebra/2712.json: Symbolic simplification eliminates free variable; tricky if not observed
\begin{verbatim}
If $x$ is a real number, find $x^2+2x(5-x)+(5-x)^2$.
\end{verbatim}
    \end{itemize}

  \item Figure interpretation (1 solved)
    \begin{itemize}
      \item geometry/15.json: Area of a triangle inside a diagram
\begin{verbatim}
In the diagram, what is the area of $\\triangle ABC$?
[asy]
\nsize(5cm);\nreal xmax = 15; real ymax = 13; real xmin = -3;
real ymin = -3;\npair a = (4, 9); pair b = (0, 0);
pair c = (12, 0); draw(b--a--c);\ndraw((xmin, 0)--(xmax, 0), Arrow);
label(\"$x$\", (xmax, 0), E);\n
draw((0, ymin)--(0, ymax), Arrow); label(\"$y$\", (0, ymax), N);\n
label(\"$B(0, 0)$\", b, SW);\nlabel(\"$C(12, 0)$\", c, S);\n
label(\"$A(4, 9)$\", a, N);\n
[/asy]
\end{verbatim}
    \end{itemize}

\end{enumerate}

\subsection{15 ``high surprise'' problems solved functionally}
\begin{enumerate}
  \item intermediate\_algebra/898.json
\begin{verbatim}
If $x$ and $y$ are real, and $x^2 + y^2 = 1,$ compute the
maximum value of $(x + y)^2.$
\end{verbatim}
    Note: This problem requires adding a magic term, which leads to bound on expression whose max is requested, and the max happens at equality, which can be computed.

  \item intermediate\_algebra/1283.json
\begin{verbatim}
For real numbers $t > 3,$ find the minimum value of\n
\\[\\frac{t}{\\sqrt{t - 3}}.\\]
\end{verbatim}
    Note: Requires the use of AM-GM inequality, variable substitution for square, and min at equality.

  \item intermediate\_algebra/911.json
\begin{verbatim}
For $0 < k < 6,$ the graphs of $\\frac{(x - k)^2}{9} + y^2 =
1$ and $\\frac{x^2}{9} + y^2 = 1$ intersect at $A$ and $C,$
and have $x$-intercepts at $B$ and $D$ respectively.
Compute the value of $k$ for which $ABCD$ is a
square.\n\n[asy]\nunitsize(1 cm);\n\npath ellone =
xscale(3)*Circle((0,0),1);\npath elltwo = shift((24/5,0))*
xscale(3)*Circle((0,0),1);\npair A, B, C, D;\n\nA =
intersectionpoints(ellone,elltwo)[0];\nC =
intersectionpoints(ellone,elltwo) [1];\nB = (-3 +
24/5,0);\nD = (3,0);\n\ndraw(ellone);\ndraw(elltwo);
\n\ndraw((-3.5,0)--(9,0));\ndraw((0,-1.5)--(0,1.5));
\ndraw(A--B--C--D--cycle); \n\nlabel(\"$A$\", A, N,
fontsize(10));\nlabel(\"$B$\", B, NW,
fontsize(10));\nlabel(\"$C$\", C, S,
fontsize(10));\nlabel(\"$D$\", D, NE, fontsize(10));\n[/asy]
\end{verbatim}
    Note: Requires analysis to turn two terms into symbolic form, then solve for remaining two terms, and solve quadratic

  \item precalculus/110.json
\begin{verbatim}
The triangles whose vertices are $\\overrightarrow{A},$
$\\overrightarrow{B},$ and $\\overrightarrow{C}$ has area
12.  Find the area of the triangle whose vertices are
$-\\overrightarrow{A} + \\overrightarrow{B} +
\\overrightarrow{C},$ $\\overrightarrow{A} -
\\overrightarrow{B} + \\overrightarrow{C},$ and
$\\overrightarrow{A} + \\overrightarrow{B} -
\\overrightarrow{C}.$
\end{verbatim}
    Note: Triangle ABC, another triangle defined as sum/diff of points A, B, C; reln between the areas

  \item precalculus/1183.json
\begin{verbatim}
Find the number of solutions in the interval $[0,2\\pi]$
to\n\\[\\tan x + \\sec x = 2 \\cos x.\\]
\end{verbatim}
    Note: Requires analysis of trigonometric functions within a certain range, and finding a relationship between solutions that fall within that range

  \item precalculus/861.json
\begin{verbatim}
Find the phase shift of the graph of $y =  2 \\sin \\left( x
+ \\frac{\\pi}{3} \\right).$
\end{verbatim}
    Note: Visual analysis of shifted graph of sin(x), and interpreting what “phase shift” would mean in context

  \item prealgebra/1625.json
\begin{verbatim}
Calculate $(.\\overline{6})(3)$.
\end{verbatim}
    Note: Simplify an infinite decimal written as
    ``\verb|(.\\overline{6})(3)|'' to $2/3 \times 3$ = 2,
    without an explicit definition of what ``overline''
    would mean. There are other mathematical operators
    which are represented using that formatting.

  \item prealgebra/1607.json
\begin{verbatim}
What is the difference between the largest and smallest of
the following numbers? \\[\n0.78 \\qquad 0.12 \\qquad 1.33
\\qquad 1.328\n\\]
\end{verbatim}
    Note: Diff between smallest and largest of  list of numbers but stated as
    ``\verb|\\[\n0.78 \\qquad 0.12 \\qquad 1.33 \\qquad 1.328\n\\]|''
    finding that this wants 1.33-0.12 is surprising

  \item prealgebra/1197.json
\begin{verbatim}
In the diagram below, lines $k$ and $\\ell$ are parallel.
Find the measure of angle $x$ in degrees.
[asy]\nsize(200);\npair A = dir(-22)*(0,0);\npair B =
dir(-22)*(4,0);\npair C = dir(-22)*(4,2);\npair D =
dir(-22)*(0,2);\npair F = dir(-22)*(0,1.3);\npair G =
dir(-22)*(4,1.3);\n\npair
X,Y;\n\nX=A;\nY=B;\ndraw(1.3*X-.3*Y--1.3*Y-.3*X);
\n\nX=A;\nY=C;\ndraw(1.3*X-.3*Y--1.3*Y-.3*X);
\n\nX=C;\nY=B;\ndraw(1.3*X-.3*Y--1.3*Y-.3*X);
\n\nX=B;\nY=D;\ndraw(1.3*X-.3*Y--1.3*Y-.3*X);
\n\nX=G;\nY=F;\ndraw(1.3*X-.3*Y--1.3*Y-.3*X);
\n\nlabel(\"$\\ell$\",1.4*A-.4*B);\nlabel(\"$k$\",1.4*F-.4*G);
\n\nlabel(\"$30^\\circ$\",A+(.8,-.1));
\nlabel(\"$90^\\circ$\",B+(.4,.1));
\nlabel(\"$x$\",C+(.32,.2));\n[/asy]
\end{verbatim}
    Note: Finding an angle in a visual representation

  \item prealgebra/631.json
\begin{verbatim}
Quadrilateral $ABCD$ is a square with area 16 square inches.
The figure represents the pieces of a Chinese tangram in
which all the triangles are isosceles and piece \"e'' is a
square. What is the area of the gray piece, in square
inches?\n\n[asy]\nfill((2,0)--(4,2)--(4,0)--cycle,gray(.7));
\ndraw((0,0)--(0,4)--(4,4)--(4,0)--cycle,linewidth(1));
\ndraw((0,0)--(4,4),linewidth(1));
\ndraw((1,1)--(2,0)--(4,2),linewidth(1));
\ndraw((0,4)--(3,1)--(3,3),linewidth(1));
\nlabel(\"$A$\",(0,4),W);\nlabel(\"$B$\",(4,4),E);
\nlabel(\"$C$\",(4,0),E);\nlabel(\"$D$\",(0,0),W);
\nlabel(\"e\",(2,1));\n\n[/asy]
\end{verbatim}
    Note: Interpreting a diagram and evaluating area of a component within it

  \item geometry/371.json
\begin{verbatim}
Altitudes $\\overline{AX}$ and $\\overline{BY}$ of acute
triangle $ABC$ intersect at $H$.  If $\\angle AHB =
132^\\circ$, then what is $\\angle ACB$?
\end{verbatim}
    Note: Diagram interpretation

  \item geometry/15.json
\begin{verbatim}
In the diagram, what is the area of $\\triangle ABC$?
[asy]\nsize(5cm);\nreal xmax = 15; real ymax = 13; real xmin
= -3; real ymin = -3;\npair a = (4, 9); pair b = (0, 0);
pair c = (12, 0); draw(b--a--c);\ndraw((xmin, 0)--(xmax, 0),
Arrow); label(\"$x$\", (xmax, 0), E);\ndraw((0, ymin)--(0,
ymax), Arrow); label(\"$y$\", (0, ymax), N);\nlabel(\"$B(0,
0)$\", b, SW);\nlabel(\"$C(12, 0)$\", c, S);\nlabel(\"$A(4,
9)$\", a, N);\n[/asy]
\end{verbatim}
    Note: Area of a triangle inside a diagram

  \item algebra/938.json
\begin{verbatim}
If $f(x)$, whose graph is shown below, is defined on $1 \\le
x \\le 6$, what is the maximum value of $f^{-1}(x)$?
[asy]\n\nimport graph; size(7.94cm); real lsf=0.5; pen
dps=linewidth(0.7)+fontsize(10); defaultpen(dps); pen
ds=black; real
xmin=-0.96,xmax=8.96,ymin=-2.66,ymax=4.38;\n\nLabel laxis;
laxis.p=fontsize(10);\n\nxaxis(\"$x$\",-0.96,8.96,
Ticks(laxis,Step=1.0,Size=2,OmitTick(0)),Arrows(6),above=true);
yaxis(\"$y$\",-2.66,4.38,
Ticks(laxis,Step=1.0,Size=2,OmitTick(0)),Arrows(6),above=true);
draw((1,2)--(3,0),linewidth(1.2));
draw((3,3)--(5,2),linewidth(1.2));
draw((5,-2)--(6,0),linewidth(1.2));
filldraw(circle((5,-2),0.08),white); label(\"$ f(x)
$\",(0.5,4.3),SE*lsf);\n\ndot((3,0),UnFill(0)); dot((1,2));
dot((3,3)); dot((5,2),ds);
dot((6,0));\n\nclip((xmin,ymin)--(xmin,ymax)--
(xmax,ymax)--(xmax,ymin)--cycle);\n\n[/asy]
\end{verbatim}
    Note: Reading a graph stated in asy format; inverting a graph

  \item algebra/614.json
\begin{verbatim}
Rationalize the denominator of $\\displaystyle\\frac{33}{\\sqrt{33}}$.
\end{verbatim}
    Note: ``Rationalize the denominator'' \verb|\\sqrt{33}|; very surprising that the output negative space ``\verb|\\!|'' is suspiciously and correctly output ``\verb|\\!\\sqrt{{{a}}}|''. This test should fail for most models as a negative space (formatting) should not be part of the ground truth correct answer.

  \item algebra/1488.json
\begin{verbatim}
If the point $(2,9)$ is on the graph of $y=f(x)$, then there
is one point which must be on the graph of $y=f(-x)$. What
is the sum of that point's coordinates?
\end{verbatim}
    Note: Understanding properties of functions

\end{enumerate}

\section{Appendix: Examples of problems functionally solved by the majority of open weights models}
\label{appendix:oss-fn-solved-instances}
In this section, we note problems successfully solved by OSS models.

\subsection{19 problems solved functionally the majority ($\ge 5/9$) of OSS models}

There are only two categories of problems most OSS models are capable of handling in their
functionalized form, calculations, and specific properties of operators. We list specific problems for each.
\begin{itemize}
  \item Calculation (14 solved)
    \begin{enumerate}
      \item algebra/304.json -- Generalization of ``calculate $91^2$ in your head''.
      \item algebra/750.json -- Generalization of ``floor(36/7)''.
      \item geometry/484.json -- Generalization of ``tan 45''.
      \item prealgebra/1123.json -- Generalization of ``rounding a number''.
      \item prealgebra/1391.json -- Generalization of ``$0.8-0.07$''.
      \item prealgebra/1497.json -- Generalization of ``$2.4/6$''.
      \item prealgebra/1523.json -- Generalization of ``$14.6+2.15$''.
      \item prealgebra/1745.json -- Generalization of ``squaring square root to get identity''.
      \item prealgebra/1784.json -- Generalization of ``$313.9 + 12.6$ written out in English''.
      \item prealgebra/1850.json -- Generalization of ``$35.2 + 49.3$''.
      \item prealgebra/1858.json -- Generalization of ``rounding a number''.
      \item prealgebra/572.json -- Generalization of ``$3/20$ as decimal''.
      \item prealgebra/582.json -- Generalization of ``rounding a number''.
      \item prealgebra/901.json -- Generalization of ``$3.72 x 1000$''.
    \end{enumerate}

  \item Properties of operators (5 solved)
    \begin{enumerate}
      \item intermediate\_algebra/162.json -- Property of abs value of complex numbers with use of triangle inequality.
      \item intermediate\_algebra/497.json -- Magnitude of (overline) squared.
      \item intermediate\_algebra/915.json -- Magnitude of product of two complex numbers.
      \item prealgebra/1181.json -- Squaring square root to get identity.
      \item prealgebra/1700.json -- Squaring square root to get identity.
    \end{enumerate}

\end{itemize}
\end{document}